\def\year{2020}\relax
\documentclass[letterpaper]{article} 
\usepackage{aaai20}  
\usepackage{times}  
\usepackage{helvet} 
\usepackage{courier}  
\usepackage[hyphens]{url}  
\usepackage{graphicx} 
\usepackage{graphicx}  
\usepackage{multirow}
\usepackage{amsmath}
\usepackage{amssymb}
\usepackage[ruled,linesnumbered]{algorithm2e}  

\DeclareMathOperator{\relu}{ReLU}
\DeclareMathOperator{\sigmoid}{\sigma}
\newcommand{\vvec}[1]{\mathbf{#1}} 
\newcommand{\transpose}{\intercal} 
\usepackage{subfigure}

\title{Cross-Modality Attention with Semantic Graph Embedding \\ for Multi-Label Classification}
\author{Renchun You,\textsuperscript{\rm 1}\footnotemark[1]
Zhiyao Guo,\textsuperscript{\rm 2}\footnotemark[1]
Lei Cui,\textsuperscript{\rm 3}\thanks{These authors share first authroship.}
Xiang Long,\textsuperscript{\rm 1}
Yingze Bao,\textsuperscript{\rm 1}
Shilei Wen\textsuperscript{\rm 1}\thanks{Corresponding author.}\\
\textsuperscript{\rm 1}Baidu VIS \\
\textsuperscript{\rm 2} Computer Science Department, Xiamen University, China\\
\textsuperscript{\rm 3} Department of Computer Science and Technology, Tsinghua University, China \\
yourenchun@baidu.com, guozhiyao45@gmail.com, cuil19@mails.tsinghua.edu.cn,  longxiang@baidu.com, \\
baoyingze@gmail.com, wenshilei@baidu.com
}
\begin{document}

\maketitle

\begin{abstract}
Multi-label image and video classification are fundamental yet challenging tasks in computer vision. The main challenges lie in capturing spatial or temporal dependencies between labels and discovering the locations of discriminative features for each class. In order to overcome these challenges, we propose to use cross-modality attention with semantic graph embedding for multi-label classification. Based on the constructed label graph, we propose an adjacency-based similarity graph embedding method to learn semantic label embeddings, which explicitly exploit label relationships. Then our novel cross-modality attention maps are generated with the guidance of learned label embeddings. Experiments on two multi-label image classification datasets (MS-COCO and NUS-WIDE) show our method outperforms other existing state-of-the-arts. In addition, we validate our method on a large multi-label video classification dataset (YouTube-8M Segments) and the evaluation results demonstrate the generalization capability of our method.
\end{abstract}

\section{Introduction} \label{sec:intro}
Multi-label image classification (MLIC) and multi-label video classification (MLVC) are important tasks in computer vision, where the goal is to predict a set of categories present in an image or a video. Compared with single-label classification (e.g. assigns one label to an image or video), multi-label classification is more useful in many applications such as internet search, security surveillance, robotics, etc. Since MLIC and MLVC are very similar tasks, in the following technical discussion we will mainly focus on MLIC, whose conclusions can be migrated to MLVC naturally.

Recently, single-label image classification has achieved great success thanks to the evolution of deep Convolutional Neural Networks (CNN) \cite{r08-he2016deep,r09-huang2017densely,r10-simonyan2014very,r11-szegedy2016rethinking}.
Single-label image classification can be naively extended to MLIC tasks by treating the problem as a series of single-label classification tasks.
However, such naive extension usually provides poor performance, since the semantic dependencies among multiple labels are ignored, which are especially important for multi-label classification.
Therefore, a number of prior works aim to capture the label relations by Recurrent Neural Networks (RNN). 
However, these methods do not model the explicit relationships between semantic labels and image regions, thus they lack the capacity of sufficient exploitation of the spatial dependency in images.

An alternative solution for MLIC is to introduce object detection techniques.
Some methods \cite{wei2014cnn,zhang2018multilabel,Hao2016Exploit} extract region proposals using extra bounding box annotations, which are much more expensive to label than simple image level annotations.
Many other methods \cite{r53-wang2017multi,r56-zhu2017learning} apply attention mechanism to automatically focus on the regions of interest.
However, the attentional regions are learned only with image-level supervision, which lacks explicit semantic guidance.

To address above issues, we argue that an effective model for multi-label classification should reach two capacities: (1) capturing semantic dependencies among multiple labels in terms of spatial context; (2) locating regions of interest with more semantic guidance.

In this paper, we propose a novel cross-modality attention network associated with graph embedding, so as to simultaneously search for discriminative regions and label spatial semantic dependencies.
Firstly, we introduce a novel Adjacency-based Similarity Graph Embedding (ASGE) method which captures the rich semantic relations between labels.
Secondly, the learned label embedding will guide the generation of attentional regions in terms of cross-modality guidance, which is referred to as Cross-modality Attention (CMA) in this paper. 
Compared with traditional self-attention methods, our attention explicitly introduces the rich label semantic relations.
Benefiting from the CMA mechanism, our attentional regions are more meaningful and discriminative. Therefore they capture more useful information while suppressing the noise or background information for classification. Furthermore, the spatial context dependencies of labels will be captured, which further improve the performance in MLIC.

\begin{figure}
    \centering
    \includegraphics[width=1.0\columnwidth]{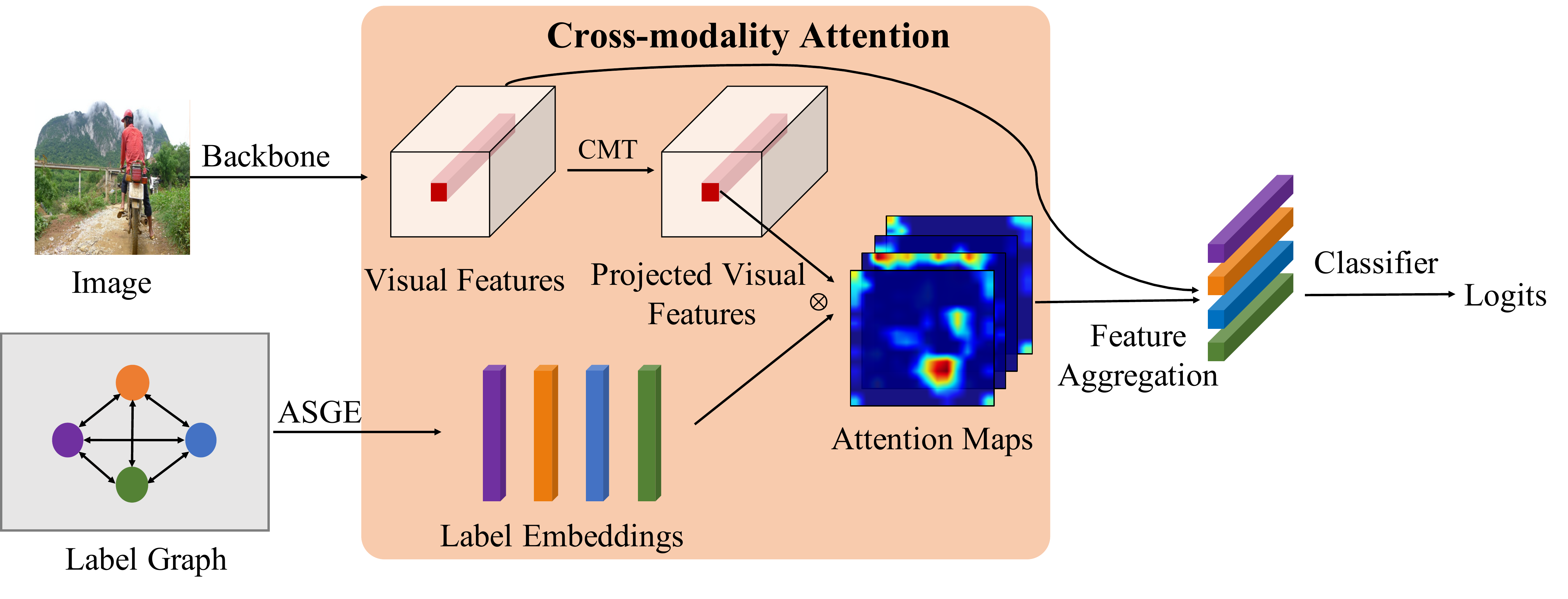}
    \caption{The overall framework of our model for MLIC task. The label embeddings are obtained by ASGE module. The visual features are first extracted by backbone network and then projected to semantic space to get projected visual features through CMT module. The learned label embeddings and projected visual features are together fed into CMA module to generate the category-wise attention maps, each of which is utilized to weightedly average the visual features and generate category-wise aggregated feature. Finally, the classifier is applied for final prediction.}
    \label{fig:imgpip}
\end{figure}

\begin{figure}
    \centering
    \includegraphics[width=1.0\columnwidth]{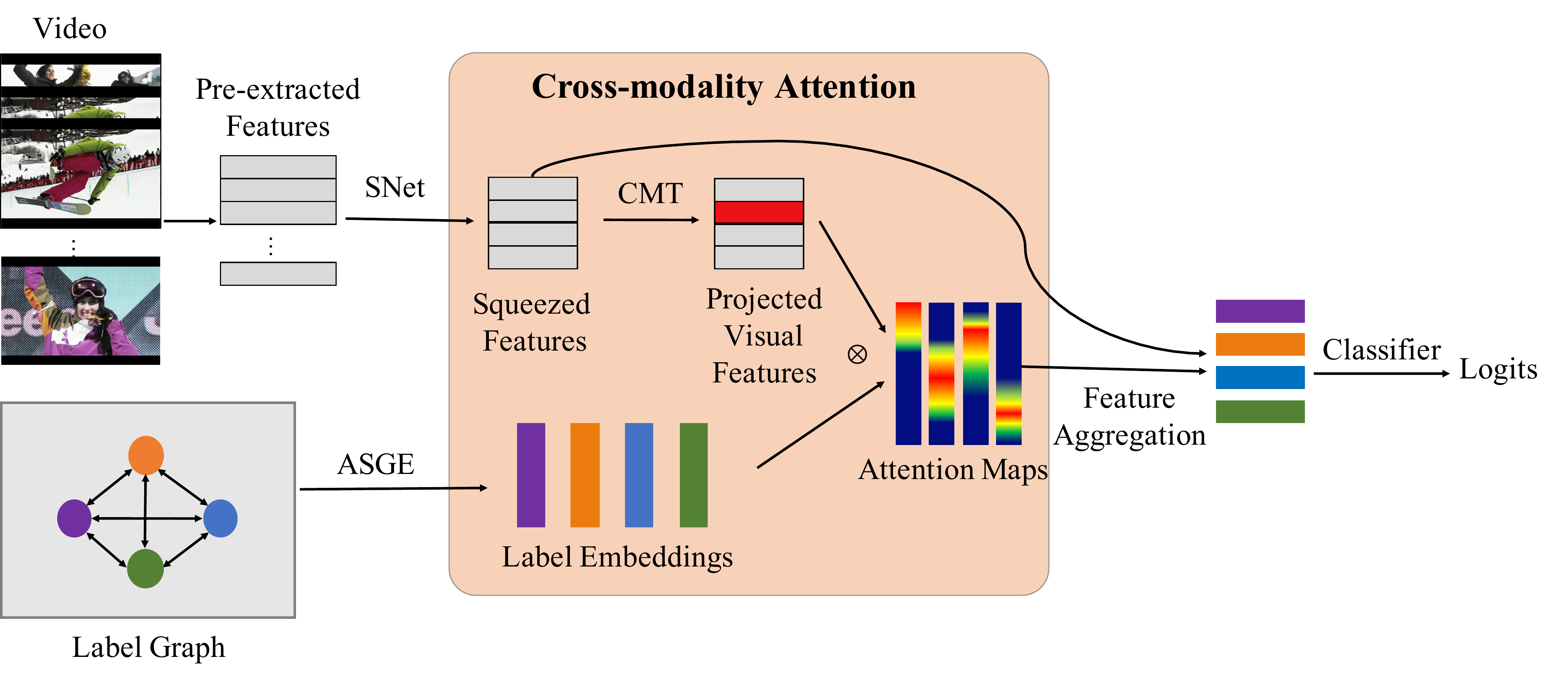}
    \caption{The overall framework of the model for MLVC task. The input is pre-extracted features instead of raw video, and the pre-extracted features are processed through SNet to extract visual features. Other parts are quite similar to that of MLIC.}
    \label{fig:vidpip}
\end{figure}

The major contributions of this paper are briefly summarized as follows:
\begin{itemize}
	\item We propose an ASGE method to learn semantic label embedding and exploit label correlations explicitly.
	
	\item We propose a novel attention paradigm, namely cross-modality attention, where the attention maps are generated by leveraging more prior semantic information, resulting in more meaningful attention maps.
	
	\item A general framework combining CMA and ASGE module, as shown in Fig.\ref{fig:imgpip} and Fig.\ref{fig:vidpip}, is proposed for multi-label classification, which can capture dependencies between spatial and semantic space and discover the location of discriminative feature effectively. We evaluate our framework on MS-COCO dataset and NUS-WIDE dataset for MLIC task, and new state-of-the-art performances are achieved on both of them. We also evaluate our proposed method on YouTube-8M dataset for MLVC, which also achieves remarkable performances. 
\end{itemize}
	    
\section{Related Works}

The task of MLIC has attracted an increasing interest recently.
The easiest way to address this problem is to treat each category independently, then the task can be directly converted into a series of binary classification tasks \cite{r24-chen2019multi}. However, such techniques are limited by without considering the relationships between labels.

Several approaches have been applied to model the correlations between labels.  \citeauthor{r25-read2011classifier}~\shortcite{r25-read2011classifier} extends the multi-label classification by training the chain binary-classifiers and introducing the correlations between labels by inputting the previously predicted labels. 
Some other works \cite{r26-li2014multi,r01-wang2016cnn,r28-chen2018order,r27-li2016conditional} formulate the task as a structural inference problem based on probabilistic graphical models. Besides, the latest work \cite{r24-chen2019multi} explores the label dependencies by graph convolutional network. However, none of aforementioned methods consider the associations between semantic labels and image contents, and the spatial contexts of images have not been sufficiently exploited. 

In MLIC task, visual concepts are highly related with local image regions. To explore information in local regions better, some works \cite{wei2014cnn,Hao2016Exploit} introduce region proposal techniques to focus on informative regions.  \citeauthor{wei2014cnn}~\shortcite{wei2014cnn} extracts an arbitrary number of object hypotheses, then inputs them into the shared CNN and aggregates the output with max pooling to obtain the ultimate multi-label predictions. \citeauthor{Hao2016Exploit}~\shortcite{Hao2016Exploit} introduces local information provided by generated proposals to  boost the discriminative power of feature extraction.
Although above methods have used region proposals to enhance feature representation, they are still limited by requiring extra object-level annotations and without considering the dependencies between objects.

Alternatively, \citeauthor{r53-wang2017multi}~\shortcite{r53-wang2017multi} discovers the attentional regions corresponding to multiple semantic labels by spatial transformer network and captures the spatial dependencies of the regions by Long Short-Term Memory (LSTM). Analogously, \citeauthor{r56-zhu2017learning}~\shortcite{r56-zhu2017learning} proposes the spatial regularization network to generate label-related attention maps and capture the latent relationships by attention maps implicitly. 
The advantage of above attention approaches is that no additional step of obtaining region proposal is needed. 
Nevertheless, the attentional regions are learned only with image-level supervision, which lacks of explicit semantic guidance. While in this paper, the semantic guidance is introduced to the generation of attention maps by leveraging label semantic embeddings, which improves the prediction performance significantly.

In this paper, the label semantic embeddings are learned by graph embedding, which is a technique aiming to learn representation of graph-structured data. The approaches of graph embedding mainly contain matrix factorization-based \cite{r39-cao2015grarep}, random walk-based \cite{r41-perozzi2014deepwalk} and neural network-based methods \cite{r45-wang2016structural,r46-zhu2018deep}. A main assumption of these approaches is the embeddings of adjacent nodes on the graph are similar, while in our task, we also require embeddings of non-adjacent nodes are mutually exclusive from each other. Therefore, we propose an ASGE method, which can further separate the embeddings of non-adjacent nodes.

MLVC is similar to MLIC, but it involves additional temporal relationships \cite{gan2015devnet,long2018multimodal,r37-campos2017skip,r38-arandjelovic2016netvlad,r35-wu2017delving,long2018attcluster}. It has been applied to many applications such as emotion recognition \cite{r29-kahou2016emonets}, human activity understanding \cite{r30-caba2015activitynet}, and event detection \cite{r31-xu2015discriminative}. In this paper, we validate our proposed method both in MLIC and MLVC task, and achieve remarkable performance.

\section{Approach} 

The overall frameworks of our approach for MLIC and MLVC are shown in Fig.\ref{fig:imgpip} and Fig.\ref{fig:vidpip} respectively. The pipeline includes several stages:
Firstly, the label graph is taken as the input of ASGE module to learn label embeddings which encode the semantic relationships between labels. 
Secondly, the learned label embeddings and visual features will be fed together into the CMA module to obtain category-wise attention maps. 
Finally, the category-wise attention maps are used to weightedly average the visual features for each category. We will describe our two key components ASGE and CMA in detail.

\subsection{Adjacency-based Similarity GE}

The relationships between labels play a crucial role in multi-label classification task as discussed in section \ref{sec:intro}. However, how to express such relationships is an open issue to be solved. Our intuition is that the co-occurrence properties between labels can be described as joint probability, which is suitable for modeling the label relationships. Nevertheless, the joint probability is easy to suffer from the influence of class imbalance. Instead, we utilize the conditional probability between labels to solve this issue, which is obtained by normalizing the joint probability through dividing by marginal probability. Based on this, it is possible to construct a label graph where the labels are nodes and the conditional probability between the labels is edge weight.  Inspired by the popular applications of graph embedding method in natural language processing (NLP) tasks, where the learned label embeddings are entered into the network as additional information, we propose a novel ASGE method to encode the label relationships.

We formally define the graph as $ \mathcal{G}=(\vvec{V},\vvec{C}) $, where $\vvec{V}=\{v_1,v_2,...,v_N \}$ represents the set of $N$ nodes and $\vvec{C}$ represents the edges. The adjacency matrix $\vvec{A} = \{A_{ij}\}_{i,j=1}^{N}$ of graph $\mathcal{G}$ contains non-negative weights associated with each edge. Specifically, $\vvec{V}$ is the set of labels and $\vvec{C}$ is the set of connections between any two labels, and the adjacency matrix $\vvec{A}$ is the conditional probability matrix by setting $A_{ij}=P(v_i/v_j)$, where $P$ is calculated through training set. Since $P(v_i|v_j)\not=P(v_j|v_i)$, namely $A_{ij}\not=A_{ji}$, in order to facilitate a better optimization, we symmetrize $\vvec{A}$ by

\begin{equation}
\vvec{A}^\prime=\frac{1}{2}\left( \vvec{A}+\vvec{A}^\transpose \right) .
\end{equation}

To capture the label correlations defined by the graph structure, we apply a neural network to map the one-hot embedding of each label $\vvec{o}_i$ to semantic embedding space and produce the label embedding
\begin{equation}
\vvec{e}_i= \Phi (\vvec{o}_i) ,
\end{equation}
where $\Phi$ denotes the neural network which consists of three fully-connected layers followed by Batch Normalization(BN) and ReLU activation. Our goal is to achieve the optimal label embedding set $\vvec{E}=\{\vvec{e}_{i}\}_{i=0}^N$ , where $\vvec{e}_{i}\in \mathbb{R}^{C_e}$. Such that $\cos(\vvec{e}_i,\vvec{e}_j)$ is close to $\vvec{A}_{ij}$ for all $i,j$, where $\cos(\vvec{e}_i,\vvec{e}_j)$ denotes the cosine similarity between $\vvec{e}_i$ and $\vvec{e}_j$. Thereby, the objective function is defined as follows:
\begin{equation} \label{eq:lossge}
\mathcal{L}_{ge} = \sum_{i=1}^{N} \sum_{j=1}^{N} \left( \frac{\vvec{e}_i^\transpose \vvec{e}_j}{ \left\| \vvec{e}_i \right\| \left\|\vvec{e}_j \right\| } -A_{ij}^{\prime} \right)^2,
\end{equation}
where $\mathcal{L}_{ge}$ denotes the loss of our graph embedding.

\subsubsection{Optimization Relaxation.}

In order to optimize Eq.\ref{eq:lossge}, the cosine similarity $\cos(\vvec{e}_i,\vvec{e}_j)$ are required to be close to the corresponding edge weight $A_{ij}$ for all $i,j$. 
However, it is hard to satisfy this strict constraint, especially when the graph is large and sparse. In order to address this problem, a hyperparameter $\alpha$ is introduced to the Eq.\ref{eq:lossge} to relax the optimization. The new objective function is as follows:
\begin{equation} \label{eq4}
\mathcal{L}_{ge} = \sum_{i=1}^{N} \sum_{j=1}^{N} \sigma_{ij}\cdot \left( \frac{\vvec{e}_i^\transpose \vvec{e}_j}{ \left\|\vvec{e}_i \right\| \left\|\vvec{e}_j \right\| } -A_{ij}^{\prime} \right)^2 ,
\end{equation}
where $\sigma_{ij}$ is an indicator function:
\begin{equation}
\sigma_{ij} = 
\begin{cases}
0, &  A_{ij}<\alpha\quad and \quad  \frac{\vvec{e}_i^\transpose \vvec{e}_j}{ \left\|\vvec{e}_i \right\| \left\|\vvec{e}_j \right\|} < \alpha\\
1, & otherwise 
\end{cases}
.
\end{equation}

By adding this relaxation, it only needs to make the embedding pairs $(e_i,e_j)$ be away instead of strictly enforcing $\cos(\vvec{e}_i,\vvec{e}_j)$ to be $A_{ij}$ when $A_{ij}<\alpha$, thus focusing more on the strong relationships between labels and reducing the difficulty of the optimization.

\subsection{CMA for Multi-label Classification}

We formally define the multi-label classification task as a mapping function  $F:\vvec{x} \rightarrow \vvec{y}$, where $\vvec{x}$ denotes a input image or video,  $\vvec{y} = [y_1, y_2, ..., y_N] $ denotes corresponding labels, $N$ is the total number of categories and $y_n \in \{0,1 \}$ denotes whether the label is assigned to the image or video.

For multi-label classification, we propose an novel attention mechanism, named cross-modality attention, which uses semantic embeddings to guide spatial or temporal integration of visual features. The semantic embeddings here are the label embedding set $\vvec{E}=\{\vvec{e}_{i}\}_{i=0}^N$ achieved by ASGE and the visual features   $\vvec{I}=\psi(\vvec{x})$ are extracted by backbone neural networks $\psi$. Note that for different tasks, we only need to apply different backbones to extract visual features, and the rest part of the framework is completely generic for both tasks. 

\subsubsection{Backbone.}
In the MLIC task, we apply ResNet-101 network to extract the last convolutional feature map as the visual features. Additionally, we use an $1 \times 1$ convolution to reduce the dimension, and obtain final visual feature map $\vvec{I} \in \mathbb{R}^{H \times W\times C}$, where $H \times W$ is the spatial resolution of the last feature map and $C$ is the number of channels. 

In the MLVC task, frame level features $\vvec{x}$ are pre-extracted by an Inception network and then processed by PCA with whitening. Considering the meaningful and discriminative information of video is derived from some pivotal frames while others may be redundant, we apply a Squeezing Network (SNet) to squeeze the temporal dimensions. The SNet is built on 4 successive $1D$ convolution and pooling layers. We can obtain the final visual feature $\vvec{I}=f_{SNet}(\vvec{x})$, where  $\vvec{I} \in \mathbb{R}^{T\times C}$ and $T$ is the temporal resolution of the final feature map and $C$ is the number of channels.

\subsubsection{Cross-Modality Attention.}
The learned label embeddings by ASGE compose a semantic embedding space, while the extracted features from CNN Backbone define a visual feature space. Our goal is to let semantic embeddings guide the generation of attention maps. However, semantic embedding space and visual feature space exist a semantic gap because of modality difference. In order to measure the compatibility between different modalities, we first learn a mapping function from the visual feature space to the semantic embedding space, then the compatibility can be measured by a cosine similarity between projected visual feature and semantic embedding, namely cross-modality attention. Formal definition is introduced as follows.

Firstly, we project the visual feature to semantic space by a Cross-Modality Transformer (CMT) module, which is built with several $1\times1$ convolution layers followed by a BN and a ReLU activation.
\begin{equation}
\vvec{I}_s=f_{cmt} (\vvec{I})
\end{equation}
where  $\vvec{I_s} \in \mathbb{R}^{M \times C_e}$ ($M=W\times H$ for image and $M=T$ for video), $f_{cmt}$  denotes the map function of the CMT module. The category-specific cross-modality attention map $z_k^{i}$ is yielded by calculating the cosine similarity between label embedding $\vvec{e}_k$ and projected visual feature vector $\vvec{I}_s^{i}$ at location $i$ of $\vvec{I}_s$:
\begin{equation} \label{eq20}
z_{k}^{i}=\relu \left( \frac{ {\vvec{I}_{s}^{i}}^\transpose \vvec{e}_k}{\left\| \vvec{I}_s^{i}  \right\| \left\|\vvec{e}_k \right\|} \right). 
\end{equation}
The category-specific attention map $z_k^{i}$ is then normalized to:
\begin{equation}  \label{eq5}
a_k^{i}= \frac{z_k^{i}}{\sum_{i=1}^M z_k^{i}  }.
\end{equation}
For each location $i$, if the CMA mechanism generates a high positive value, it can be interpreted as the location $i$ is highly semantic related to label embedding $k$ or relative more important than other locations, thus the model needs to focus on location $i$ when considering category $k$ . Then the category-specific cross-modality attention map is used to weightedly average the visual feature vectors for each category:
\begin{equation}
\vvec{h}_k=\sum_{i=1}^M \alpha_k^{i} \vvec{I}^{i},
\end{equation}
where $\vvec{h}_k$ is the final feature vector for label $k$. Then $\vvec{h}_k$ is fed into the fully-connected layers for estimating probability of category $k$:
 \begin{equation} \label{eq7}
 y_k^*=\sigmoid(\vvec{w_k}^{\transpose} \vvec{h}_k+b),
 \end{equation}
where $\vvec{w_k} \in \mathbb{R}^{C}$ and $b$ are learnable parameters. $y_k^*$ is the predicted probability for label $k$.
For convenience, we denote the calculation of whole CMA module as $y_k^* = f_{cma}(I, E)$.

Compared with general single attention map method, where the attention map is shared by all categories, our CMA module benefits in two ways: 
Firstly, our category-wise attention map is related to image regions corresponding to category $k$, thus better learn category-related regions. 
Secondly, with the guidance of label semantic embeddings, the discovered attentional regions can be better match with the annotated semantic labels.

\begin{figure}
    \centering
    \includegraphics[width=1.0\columnwidth]{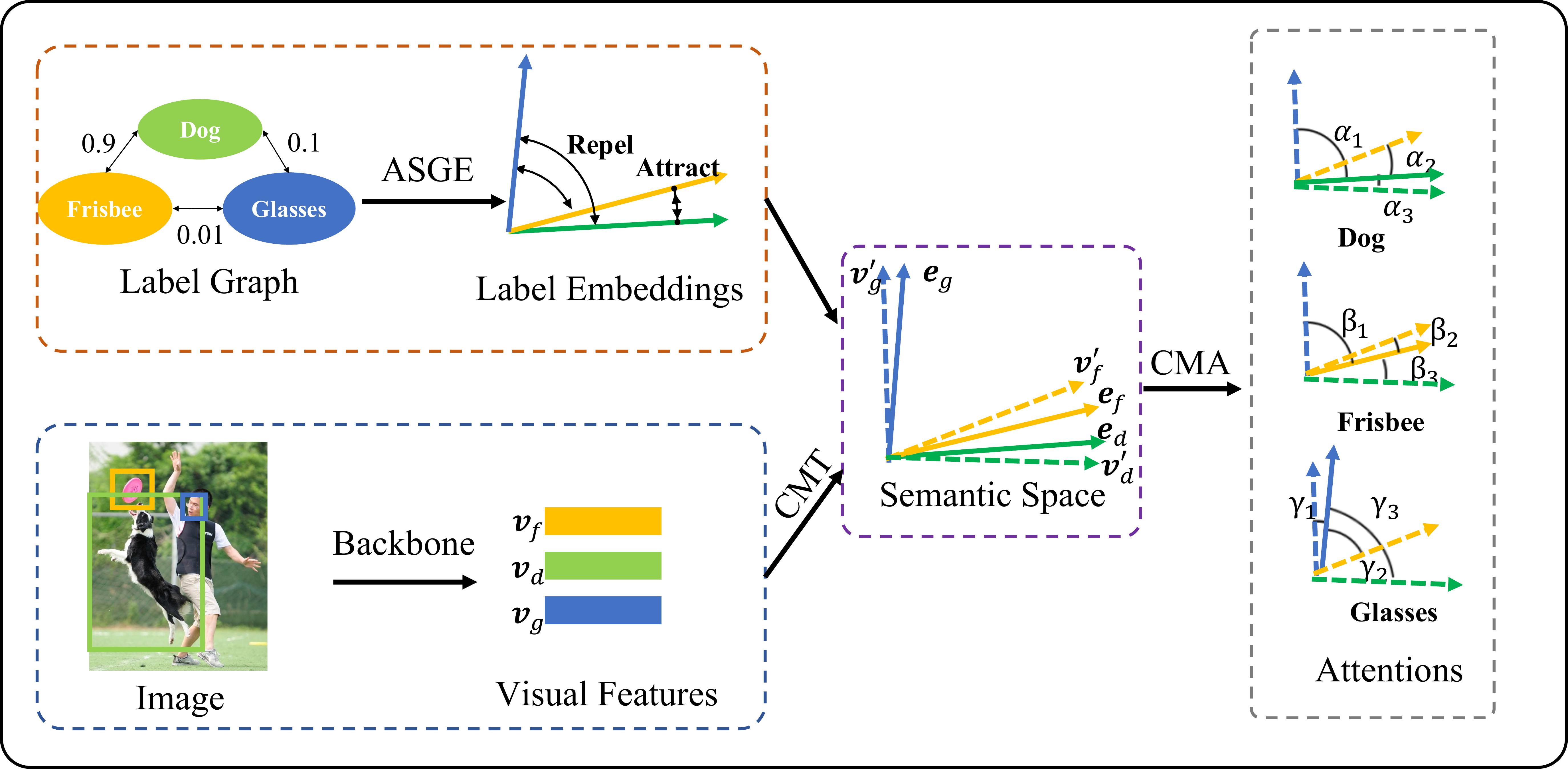}
    \caption{Illustration of latent spatial dependency. The different colors indicate different categories. The solid arrows represent the learned label embeddings, expressed as $\vvec{e}$, while the dotted arrows represent the projected visual features through CMT module, expressed as $\vvec{v}'$. The angles between label embeddings and projected visual features, namely $\alpha$, $\beta$ and $\gamma$, represent the category-wise attention scores. For a detailed discussion, refer to section 3.2.}
    \label{fig-intro}
\end{figure}

The potential advantage of our framework is to capture the latent spatial dependency, which is helpful for visual ambiguous labels. As shown in Fig.\ref{fig-intro}, we consider \textit{frisbee} as an example to explain the spatial dependency. Firstly, The ASGE module learns label embeddings through the label graph, which encodes the label relationships.  Since the \textit{dog} and \textit{frisbee} are often co-exist while \textit{glasses} not, therefore the label embeddings of \textit{dog} and \textit{frisbee} are close to each other and far away from \textit{glasses}'s, namely $\vvec{e}_d\approx{\vvec{e}_f}\not=\vvec{e}_g$. The optimization procedure during training will enforce the cosine similarity between visual feature and the corresponding label embedding become high, in other words, the $\cos(\vvec{e}_d,\vvec{v}_d'), \cos(\vvec{e}_g,\vvec{v}_g')$ and $\cos(\vvec{e}_f,\vvec{v}_f')$ will be large. Since $\vvec{e}_d\approx \vvec{e}_f\not=\vvec{e}_g$, $\cos(\vvec{e}_f,\vvec{v}_d')$ will also be large, while $\cos(\vvec{e}_f,\vvec{v}_g')$ will be small. And the final feature representation of \textit{frisbee} is $\vvec{h}_f=\beta_1 \vvec{v}_g+\beta_2 \vvec{v}_f+\beta_3 \vvec{v}_d$,where $\beta_1=\cos(\vvec{e}_f, \vvec{v}_g')$, $\beta_2=\cos(\vvec{e}_f,\vvec{v}_f')$, $\beta_3=\cos(\vvec{e}_f,\vvec{v}_d')$. Thus, the recognition of \textit{frisbee} is depending on the semantic related label \textit{dog} and not related to label \textit{glasses}, indicating that our model is capable of capturing spatial dependencies. Specially, considering that the \textit{frisbee} is a hard case to be recognized, $\beta_2$ will be small. Fortunately, the $\beta_3$ may still be large, so the visual information of \textit{dog } will be a helpful context to aid in the recognition of label \textit{frisbee}.

\subsubsection{Multi-Scale CMA.}
Single-scale feature representation may not be sufficient for multiple objects from different scales. It is noteworthy that the calculation of attention involves the label embedding slides densely over all locations of the feature map, in other words, the spatial resolution of feature map may effect on attention result. Our intuition is that the low-resolution feature maps have more representational capacity for small objects while high-resolution is opposite.
The design of CMA mechanism makes it can be naturally applied to multi-scale feature maps via a score fusion strategy. Specially, we extract a set of feature maps $\{I_1, I_2, ..., I_L\}$ and the final predicted probability of multi-scale CMA is 
\begin{equation}
y_k^*=\frac{1}{L}\sum_{l=1}^L f_{cma}(I_l, E).
\end{equation}

\subsubsection{Training Loss.}

Finally we define our object function for multi-label classification as follows
\begin{equation}
L(\theta ) =-\sum_{k=1}^{N} w_k \left[ y_k \cdot log(y_k^*) \notag + (1- y_k) \cdot (1-log(y_k^*)) \right]
\end{equation}

\begin{equation} \label{eq:wk}
w_k= y_k \cdot e^{\beta (1-p_k)}+ (1-y_k)\cdot e^{\beta p_k},
\end{equation}
Where $w_k$ is used to alleviate the class imbalance, $\beta$ is a hyperparameter and $p_k$ is the ratio of label $k$ in the training set.

\section{Experiments} \label{sec:exp}

To assess our model, we perform experiments on two benchmark multi-label image recognition datasets (MS-COCO \cite{r07-lin2014microsoft} and NUS-WIDE \cite{r21-chua2009nus}) . We also validate the effectiveness of our model on one multi-label video recognition dataset (YouTube-8M Segments) , and the results demonstrate the extensibility of our method. In this section, we will introduce the results on MLIC and MLVC respectively.

\subsection{Multi-label Image Classification}

\subsubsection{Implementation Details.}

In ASGE module, the dimensions of the three hidden layers and label embeddings are all set as 256. The optimization relaxation is not applied here since the label graph is relatively small. The optimizer is Stochastic Gradient Descent (SGD) with momentum 0.9 and the initial learning rate is 0.01.  In the classification part, the input image is randomly cropped and resized to $448 \times 448$ with random horizontal flip for augmentation. The batch size is set as 64. The optimizer is SGD with momentum 0.9. Weight decay is $10^{-5}$. The initial learning rate is 0.01 and decays by a factor 10 every 30 epochs. And the hyperparameter $\beta$ in the Eq.\ref{eq:wk} is 0. in MS-COCO dataset and 0.4 in NUS-WIDE dataset.
Based on this setup, we implement two models: CMA and Multi-Scale CMA(MS-CMA). The MS-CMA model uses three scale features $I_1 \in \mathbb{R}^{28 \times 28 \times 1024}$, $I_2 \in \mathbb{R}^{14 \times 14 \times 2048}$  from ResNet-101 backbone and $I_3 \in \mathbb{R}^{7 \times 7 \times 512}$  obtained by applying a residual block on $I_2$. While the CMA model only uses $I_2 $ .

\subsubsection{Evaluation Metrics.}

We use the same evaluation metrics as other works \cite{r53-wang2017multi}, which are the per-category and overall metrics: precision (CP and OP), recall (CR and OR) and F1 (CF1 and OF1). In addition, we also calculate the mean average precision (mAP), which is relatively more important than other metrics, and we mainly focus on the performance of mAP.

\subsubsection{Results on MS-COCO Dataset.}

The MS-COCO dataset is widely used in MLIC task. It contains 122,218 images with 80 labels and almost 2.9 labels per image. We divide the dataset into two parts: 82,081 images for training and 40,137 images for testing, according to the officially provided division criteria. 

We compare with the currently published state-of-the-art methods, including CNN-RNN\cite{r01-wang2016cnn}, RNN-Attention\cite{r53-wang2017multi}, Order-Free RNN\cite{r28-chen2018order}, ML-ZSL \cite{r55-lee2018multi}, SRN \cite{r56-zhu2017learning} and Multi-Evidence \cite{r57-ge2018multi}. Besides, we run the source code released by ML-GCN\cite{r24-chen2019multi} to train and get the results for comparison. The quantitative results of CMA and MS-CMA model are shown in Table \ref{tab1}. Our two models both perform better than the state-of-the-art methods over almost all metrics. Specially, our MS-CMA model achieves better performance than the CMA model, demonstrating the multi-scale attentions yield performance improvement. 

\begin{table*}[t]
	\centering
	\resizebox{0.98\textwidth}{!}{
		\begin{tabular}{c|c|c|c|c|c|c|c|c|c|c|c|c|c}
			\hline
			\multirow{2}{*}{Methods} 
			& \multicolumn{7}{c|}{All} & \multicolumn{6}{c}{Top-3} \\ 
			\cline{2-14}
			& mAP & CP & CR & CF1 & OP & OR & OF1 & CP & CR & CF1 & OP & OR & OF1 \\ 
			\hline
			\hline
			CNN-RNN ~\shortcite{r01-wang2016cnn} 
			& 61.2 & - & - & - & - & - & - & 66.0 & 55.6 & 60.4 & 69.2 & 66.4 & 67.8 \\ 
			CNN-LSEP ~\shortcite{r17-li2017improving}
			& - & 73.5 & 56.4 & 62.9 & 76.3 & 61.8 & 68.3 & - & - & - & - & - & - \\ 
			CNN-SREL-RNN ~\shortcite{r58-liu2017semantic}
			& - & 67.4 & 59.8 & 63.4 & 76.6 & 68.7 & 72.4 & - & - & - & - & - & - \\ 
			RNN-Attention ~\shortcite{r53-wang2017multi}
			& - & - & - & - & - & - & - & 79.1 & 58.7 & 67.4 & 84.0 & 63.0 & 72.0 \\ 
			Order-Free RNN ~\shortcite{r28-chen2018order}
			& - & - & - & - & - & - & - & 71.6 & 54.8 & 62.1 & 74.2 & 62.2 & 67.7 \\ 
			ML-ZSL ~\shortcite{r55-lee2018multi}
			& - & - & - & - & - & - & - & 74.1 & 64.5 & 69.0 & - & - & - \\ 
			SRN ~\shortcite{r56-zhu2017learning}
			& 77.1 & 81.6 & 65.4 & 71.2 & 82.7 & 69.9 & 75.8 & 85.2 & 58.8 & 67.4 & 87.4 & 62.5 & 72.9 \\ 
			S-CLs  ~\shortcite{r59-liu2018multi} 
			& 74.6 & - & - & 69.2 & - & - & 74.0 & - & - & 66.8 & - & - & 72.7 \\ 
			Multi-Evidence ~\shortcite{r57-ge2018multi}
			& - & 80.4 & 70.2 & 74.9 & \underline{85.2} & 72.5 & 78.4 & 84.5 & 62.2 & 70.6 & 89.1 & 64.3 & 74.7 \\ 
			ML-GCN  ~\shortcite{r24-chen2019multi}
			& 82.4 & 82.1 & \underline{73.1} & 77.3 & 83.7 & \underline{76.3} & 79.9 & \underline{87.2} & 64.6 & 74.2 & 89.1 & 66.7 & 76.3 \\ 
            \hline
			\textbf{CMA} 
			& \underline{83.4} & \textbf{83.4} & 72.9 & \underline{77.8} & \textbf{86.8} & 76.3 & \underline{80.9} & 86.7 & \underline{64.9} & \underline{74.3} & \textbf{90.9} & \underline{67.2} & \textbf{77.2} \\ 
			\textbf{MS-CMA} 
			& \textbf{83.8} & \underline{82.9} & \textbf{74.4} & \textbf{78.4} & 84.4 &\textbf{ 77.9} & \textbf{81.0} & \textbf{88.2} & \textbf{65.0} & \textbf{74.9} & \underline{90.2} &\textbf{ 67.4} & \underline{77.1} \\
			\hline
		\end{tabular}
	}
	\caption{Comparisons with state-of-the-art methods on the MS-COCO dataset. We report two our proposed model: \textbf{CMA} and \textbf{MS-CMA}. The bold numbers indicate the best results in different metrics, while the underlined numbers indicate the suboptimal results.}
	\label{tab1}
\end{table*}

\subsubsection{Results on NUS-WIDE Dataset.}

The NUS-WIDE is a web dataset including 269,648 images and 5018 labels from the Flickr. After removing the noise and the rare labels, there are 1000 categories left. The images are further manually annotated into 81 concepts with 2.4 concepts per image on average. We follow the split used in \cite{r59-liu2018multi}, i.e. 150,000 images for training and 59,347 for testing after removing the images without any labels.

In this dataset, we compare with the current state-of-the-art models, including CNN-RNN \cite{r01-wang2016cnn} , CNN-SREL-RNN \cite{r58-liu2017semantic} , CNN-LSEP \cite{r17-li2017improving}, Order-Free RNN\cite{r28-chen2018order}, ML-ZSL \cite{r55-lee2018multi}, S-CLs\cite{r59-liu2018multi}, Attention transfer\cite{r60-zagoruyko2016paying} and FitsNet\cite{r61-romero2014fitnets}. 

\begin{table}[t]
	\centering
	\resizebox{0.47\textwidth}{!}{
		\begin{tabular}{c|c|c|c|c|c}
			\hline
			\multirow{2}{*}{Methods} 
			& \multicolumn{3}{c|}{All} & \multicolumn{2}{c}{Top-3} \\ 
			\cline{2-6}
			& mAP  & CF1 & OF1 & CF1 & OF1 \\ 
			\hline
			\hline
			CNN-RNN ~\shortcite{r01-wang2016cnn}
			&56.1&-&-&34.7&55.2\\ 
			CNN-LSEP ~\shortcite{r17-li2017improving}
			&-&52.9&70.8&-&-\\ 
			CNN-SREL-RNN ~\shortcite{r58-liu2017semantic}
			&-&52.7&70.9&-&-\\ 
			Order-Free RNN ~\shortcite{r28-chen2018order}
			&-&-&-&54.7&\underline{70.2 }\\ 
			ML-ZSL ~\shortcite{r55-lee2018multi}
			&-&-&-&45.7&-\\ 
			Attention transfer  ~\shortcite{r60-zagoruyko2016paying}
			&57.6&55.2&70.3&51.7&68.8\\ 
			FitsNet ~\shortcite{r61-romero2014fitnets}
			&57.4&54.9&70.4&51.4&68.6\\ 
			S-CLs ~\shortcite{r59-liu2018multi}
			&60.1&58.7&\underline{73.7}&53.8&\textbf{71.1} \\ 
            \hline
            \textbf{CMA} &\underline{60.8}&\underline{60.4}&\underline{73.7 }&\underline{55.5}&70.0\\ 
			\textbf{MS-CMA} 
			&\textbf{61.4}&\textbf{60.5}&\textbf{73.8}&\textbf{55.7}&69.5\\
			\hline
		\end{tabular}
	}
	\caption{Comparisons with state-of-the-art methods on the NUS-WIDE dataset.}
	\label{tab2}
\end{table}

The quantitative results are shown in the Table \ref{tab2}. 
The comparison results are similar to MS-COCO's. Our CMA and MS-CMA perform better than state-of-the-art methods on most metrics. The mAP metric of our MS-CMA, which is mostly concerned, exceeds the previous state-of-the-art result by $1.3\%$. 
 We observe that the average edge weight per label of NUS-WIDE is 3.3, while that of MS-COCO is 3.9. This shows that the label graph of MS-COCO is denser than NUS-WIDE. And the performance gain of NUS-WIDE is slightly less obvious than that of MS-COCO. These observations indicate that richer label relationships may bring performance improvement.

\subsubsection{Ablation Study.}

In this section, we expect to answer the following questions:

\begin{itemize}
	\item Comparing with the backbone (ResNet-101) model, does our CMA model improve significantly?
	\item Does our proposed CMA mechanism have an advantage over the general self-attention methods?
    \item Can the CMA extend to multi-scale and bring performance improvement? 
    \item Is our ASGE more advantageous than other embedding methods, e.g. Word2vec?
\end{itemize}{}

To answer these questions, we conduct some ablation studies on the MS-COCO dataset, as shown in table \ref{tab3}.  
Firstly, we investigate how CMA contributes to mAP. It is obvious to see that the vanilla ResNet-101 achieves $79.9\%$ mAP, while increases to $83.4\%$ when CMA module is added. This result shows the significant effectiveness of the CMA mechanism. 
Secondly, we implement a general self-attention method by replacing  Eq.\ref{eq20}  with  $z^{i}=\sigmoid(f_{conv}(I^i))$, where $f_{conv}$ denotes the map function of $1\times1$ convolution layer. 
Our CMA mechanism performs better than general self-attention mechanism by achieving $2.3\%$ mAP improvement, which indicates that the label semantic embeddings guided attention mechanism is superior to the general self-attention due to introducing much more prior information. 
Thirdly, expanding our CMA mechanism to multiple scales could obtain about $0.4\%$ improvement. This result demonstrates that our attention mechanism is well adapted for multi-scale features. 
Finally, we compared our ASGE with other embedding methods, in this paper, we take Word2vec as an exmaple, which is a group of related models used to produce word embeddings. Specially, we view the label set in each image as a single sentence, and the window size in Word2vec is set as the length of longest sentence to eliminate the influence of label order. The experiment results show our ASGE based MS-CMA performs better than Word2vec based MS-CMA (represented as W2V-MS-CMA) by $1.3\%$ improvement. In our ASGE, label relationships are explicitly represented by adjacency matrix which is treated as a direct optimization target. Instead, Word2vec implicitly encodes label relationships in a data-driven manner without directly optimizing label relationships. Therefore, our ASGE will capture label relationships much better.

\begin{table}[t]
	\centering
	\begin{tabular}{c|c}
		\hline
		Methods & mAP \\ 
		\hline
		\hline
        ResNet-101 ~\shortcite{r08-he2016deep}& 79.9 \\
		Self-attention & 81.1 \\
        W2V-MS-CMA & 82.5 \\
        \hline
		\textbf{CMA} & 83.4 \\
		\textbf{MS-CMA} & 83.8 \\
		\hline
	\end{tabular}
	\caption{Comparison of mAP with several models on the MS-COCO dataset.}
	\label{tab3}
\end{table}

\subsubsection{Visualization and Analysis.}

In this section, we visualize the learned attention maps to illustrate the ability of exploiting discriminative or meaningful regions and capturing the spatial semantic dependencies.

\begin{figure}[t!]
    \centering
    \includegraphics[width=1.0\columnwidth]{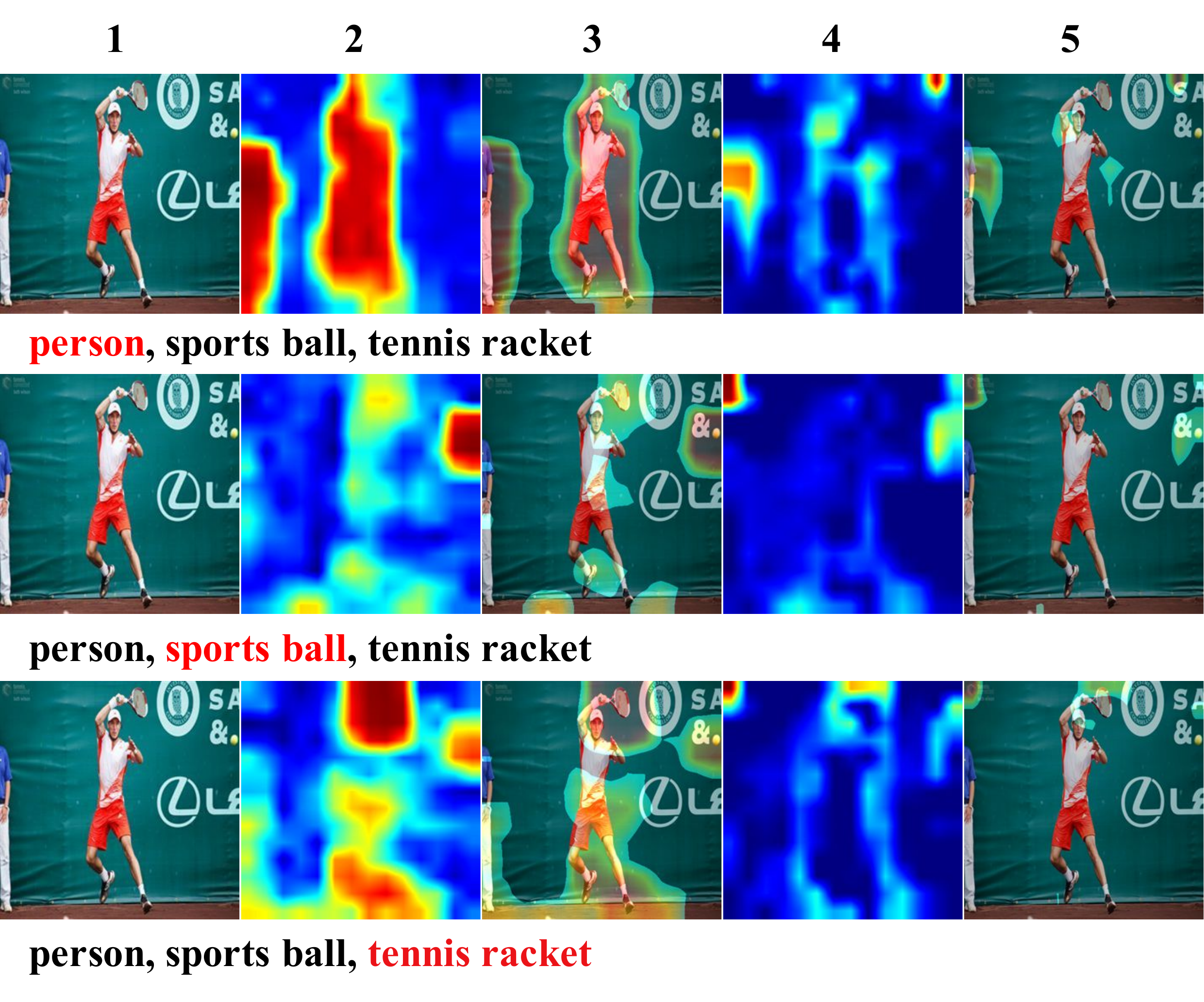}
    \caption{The visualization of attention maps. The first column: original image, second and fourth column: attention maps of MS-CMA and self-attention respectively, third and fifth column: attention maps projected on the original image of MS-CMA and self-attention respectively.}
    \label{fig5}
\end{figure}

We show the attention visualization examples in Fig.\ref{fig5}.
The three rows show the category-wise attention maps generated by CMA model and general self-attention respectively.  It is observed that the CMA model concentrates more on semantic regions and has stronger response than general self-attention, thus it is capable of exploiting more discriminative and meaningful information. Besides, our CMA mechanism has the ability of capturing the spatial semantic dependencies, especially for the indiscernible or small objects occur in the image, e.g. attention of sports ball also pays attention to tennis racket due to their semantic similarity. It's quite helpful because these objects need richer contextual cues to help recognition.
\subsection{Multi-label Video Classification}

\subsubsection{Implementation Details.}

For the training of ASGE, we apply optimization relaxation and set $\alpha=0.1$. Other settings are same as described in MLIC task. For the training of classification, the initial learning rate is 0.0002 and decay each $2\times 10^6$ samples with momentum 0.8 . The hyperparameter $\beta$ in  the Eq.12  is 0. The optimizer is SGD with momentum 0.9. The batch size is 256. In this task, we only implemented a single scale CMA model.

\subsubsection{Evaluation Metrics.}

In the MLVC task, we use several metrics to evaluate our model, including Global Average Precision (GAP)\cite{Shin2018Approach}, Average Hit Rate (Avg Hit@1), Precision at Equal Recall Rate (PERR) and Mean Average Precision (mAP)\cite{r22-abu2016youtube}.

\subsubsection{Results on YouTube-8M Segments Dataset.}

In the MLVC task, we verify the effectiveness of our model on the YouTube-8M Segments dataset, which is an extension of the YouTube-8M dataset\cite{r22-abu2016youtube}.

\begin{table}
	\centering
	\resizebox{0.47\textwidth}{!}{
		\begin{tabular}{c|c|c|c|c}
			\hline
			Methods & Avg Hit@1 & Avg PERR & mAP & GAP \\ 
			\hline
			\hline
            Self-attention-FC & 85.1 & 79.1 & 49.6 & 81.8 \\
            Self-attention-SNet & 86.1 & 80.2 & 53.2 & 83.3 \\ 
            \hline
            \textbf{CMA-FC} & 85.4 & 79.5 & 50.7 & 82.3 \\
			\textbf{CMA-SNet} & \textbf{86.7} & \textbf{81.0} & \textbf{55.8} & \textbf{84.1} \\ 
			\hline
		\end{tabular}
	}
	\caption{The comparison between self-attention model and ours CMA model on the YouTube-8M Segments dataset.}
	\label{tab5}
\end{table}

In our experiment, we only use frame-level image features, while the state-of-the-art methods use additional audio features and most are built on model ensemble, which is unfair to compare with. For this reason, we compare our CMA model with general self-attention model to validate the effectiveness in MLVC task. Besides, in order to explore the impact of backbone network on CMA mechanism, we implement the SNet-based and FC-based (two fully connected layers) models. The quantitative results are shown in the Table \ref{tab5}. It can be found that all of our metrics are better than the self-attention model. 
It seems that the improvements compared with self-attention model are not so significant as that of MLIC, but considering the input of our model is fixed pre-extracted features and the models only differ in attention mechanism, the performance gains are quite remarkable.

\subsubsection{Visualization and Analysis.}

\begin{figure}
    \centering
    \includegraphics[width=1.0\columnwidth]{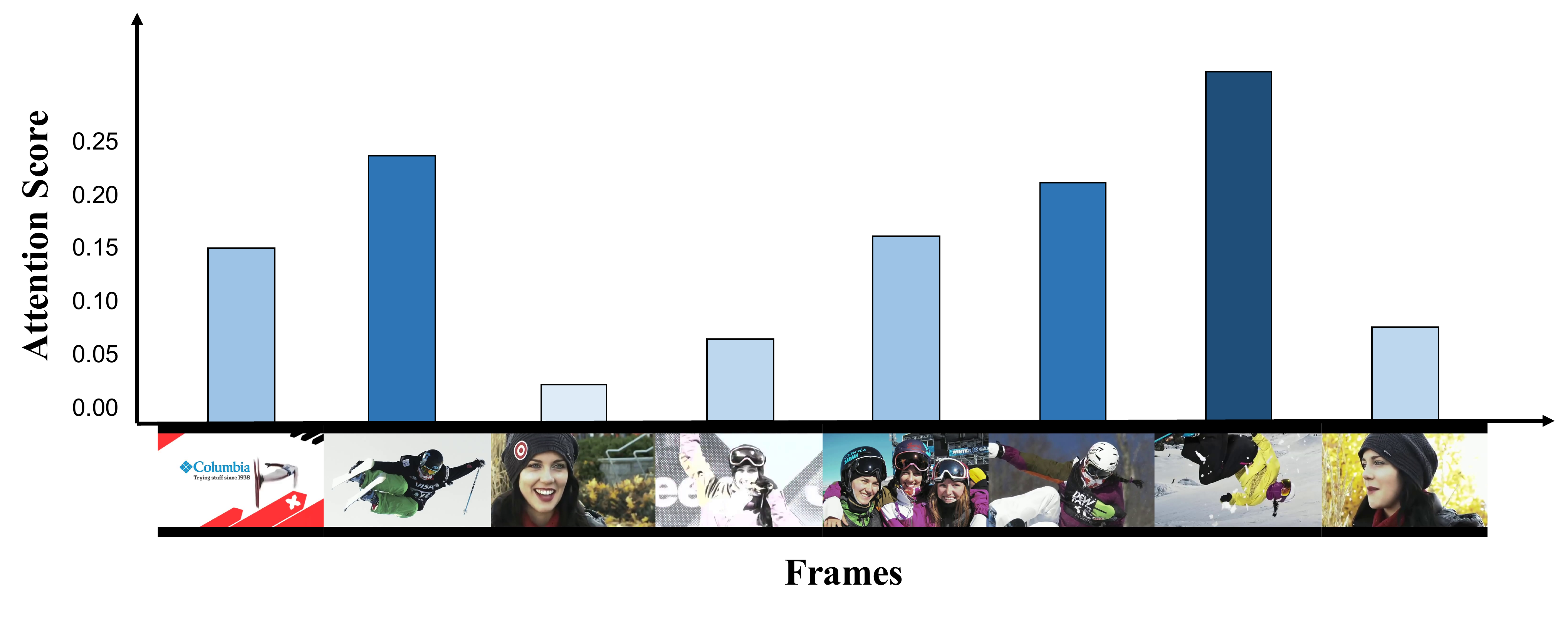}
    \caption{The attention scores for each frame. The label is \textit{skiing}.}
    \label{fig6}
\end{figure}

We also present the visualization results of attention scores for a video in Fig.\ref{fig6}. The label of the video is skiing, and our CMA model pays more attention to skiing-related frames while partly ignores the redundant frames, suggesting that our attention mechanism is capable of locating attentional frames and demonstrating the effectiveness of our model more intuitively.

\section{Conclusion}

In this paper, we propose a novel cross-modality attention mechanism with semantic graph embedding for both MLIC and MLVC task. The proposed method can  effectively discover semantic location with rich discriminative features and capture the spatial or temporal dependencies between labels. The extensive evaluations on two MLIC datasets MS-COCO and NUS-WIDE show our method outperforms state-of-the-arts. In addition, we conducted expriments on MLVC datasset YouTube-8M Segments and achieve excellent performance, which validate the strong generalization of our method.

\section{Acknowledgement}

This work was supported by the National Natural Science Foundation of China(61671397). We thank all anonymous reviewers for their constructive comments.

{
	\bibliographystyle{aaai}
	\fontsize{9.5pt}{10.5pt} \selectfont
	\bibliography{2078-references}
}

\end{document}